\documentclass[final]{cvpr}

\usepackage{times}
\usepackage{epsfig}
\usepackage{graphicx}
\usepackage{amsmath}
\usepackage{amssymb}

\usepackage{url}
\usepackage{amsfonts}
\usepackage{booktabs}
\usepackage{mathrsfs}
\usepackage{amsthm}
\usepackage{multirow}
\usepackage{cuted}
\usepackage{subfiles}
\usepackage{color}
\usepackage{threeparttable}
\usepackage[pagebackref=true,breaklinks=true,colorlinks,bookmarks=false]{hyperref}

\newcommand{\wwj}{\textcolor{black}}

\begin{document}

\title{HLA-Face: Joint High-Low Adaptation for Low Light Face Detection}

\author{Wenjing Wang, Wenhan Yang, Jiaying Liu\thanks{Corresponding author. This work was supported by the National Key Research and Development Program of China under Grant No. 2018AAA0102702 and the Fundamental Research Funds for the Central Universities and the National Natural Science Foundation of China under Contract No.61772043 and a research achievement of Key Laboratory of Science, Techonology and Standard in Press Industry (Key Laboratory of Intelligent Press Media Technology).}\\
Wangxuan Institute of Computer Technology, Peking University, Beijing, China\\
}

\maketitle

\begin{abstract}
\wwj{Face detection in low light scenarios is challenging but vital to many practical applications, e.g., surveillance video, autonomous driving at night.
Most existing face detectors heavily rely on extensive annotations, while collecting data is time-consuming and laborious.
To reduce the burden of building new datasets for low light conditions, we make full use of existing normal light data and explore how to adapt face detectors from normal light to low light.}
\wwj{The challenge of this task is that the gap between normal and low light is too huge and complex for both pixel-level and object-level.
Therefore, most existing low-light enhancement and adaptation methods do not achieve desirable performance.
To address the issue, we propose a joint High-Low Adaptation (HLA) framework.
Through a bidirectional low-level adaptation and multi-task high-level adaptation scheme, our HLA-Face outperforms state-of-the-art methods even without using dark face labels for training.
Our project is publicly available at: \url{https://daooshee.github.io/HLA-Face-Website/}
}
\end{abstract}


\section{Introduction}
\label{sec:introduction}

\begin{figure}[t]
    \centering
  \includegraphics[width=\linewidth]{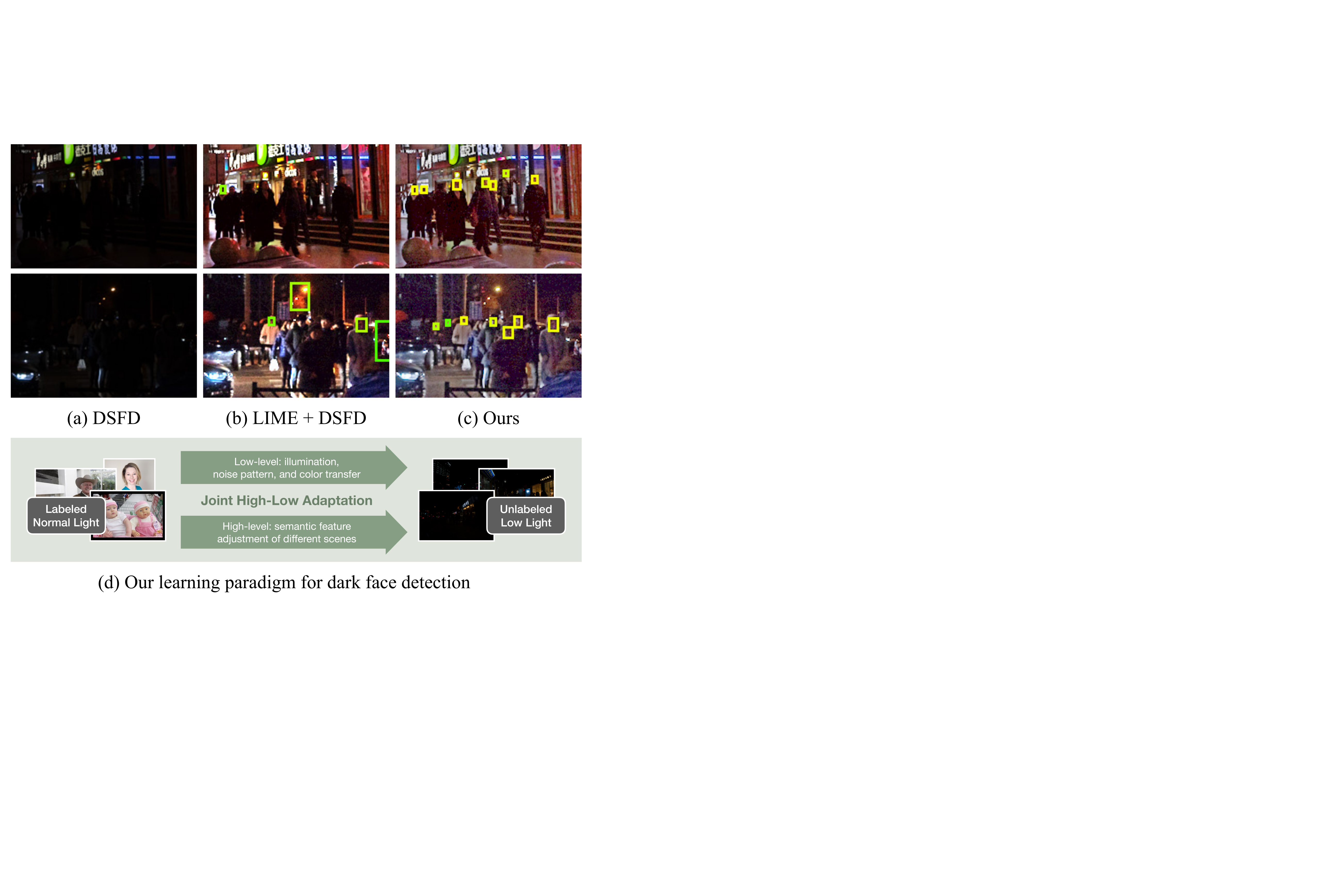}
  \caption{Dark face detection visual results and our learning paradigm. Compared with the result of DSFD~\cite{DSFD} on original low light images and the enhanced version by LIME~\cite{Enhance_LIME}, our method can better recognize the faces in dark scenarios.}
    \label{fig:teasor}
\end{figure}

Face detection is fundamental for many vision tasks, and has been widely used in a variety of practical applications, such as intelligent surveillance for smart city, face unlock, and beauty filters in mobile phones.
Over the past decades, extensive researches have made great progress in face detection.
However, face detection under adverse illumination conditions is still challenging.
Images captured without insufficient illumination suffer from a series of degradations, \eg, low visibility, intensive noise, and color cast.
These degradations can not only affect the human visual quality, but also worsen the performance of machine vision tasks, which may cause  potential risks in surveillance video analysis and nighttime autonomous driving.
In Fig.~\ref{fig:teasor}~(a), the state-of-the-art face detector DSFD~\cite{DSFD} can hardly detect faces under insufficient illumination, in direct contrast to its over 90\% precision on WIDER FACE~\cite{WIDERFACE}.

\wwj{To promote the research of low light face detection, a large scale benchmark DARK FACE~\cite{DARKFACE} is constructed.
The emergence of dark face data gives birth to a number of dark face detection researches~\cite{TMM20}.
However, existing methods are dependent on extensive annotations, therefore have poor robustness and scalability.}

In this paper, based on the benchmarking platform provided by DARK FACE, we explore how to adapt normal light face detection models to low light scenarios without the requirement of dark face annotations. 
We find that there are two levels of gaps between normal light and low light.
One is the gap in pixel-level appearance, such as the insufficient illumination, camera noise, and color bias.
The other is the object-level semantic differences between normal and low light scenes, including but not limited to the existence of street lights, vehicle headlights, and advertisement boards. 
Traditional low light enhancement methods~\cite{Enhance_LIME,Enhance_KinD} are designed for improving visual quality, therefore cannot fill the semantic gap, as shown in Fig.~\ref{fig:teasor}~(b).
Typical adaptation methods~\cite{DA_SaitoUHS19,DA_Diversify_CVPR19} are mainly designed for the scenario where the two domains share the same scene, such as adapting from Cityscapes~\cite{Cityscapes} to Foggy Cityscapes~\cite{FoggyCityscapes}.
But for our task, the domain gap is more huge, raising a more difficult challenge for adaptation.

To adapt from normal light to low light, we propose a High-Low Adaptation Face detection framework (HLA-Face).
We consider joint low-level and high-level adaptation. 
Specifically, for low-level adaptation, typical methods either brighten the dark image or darken the bright image.
However, due to the huge domain gap, they do not achieve desirable performance.
\wwj{Instead of unidirectional low-to-normal or normal-to-low translation, we bidirectionally make two domains each take a step towards each other.
By brightening the low light images and distorting the normal light images, we build intermediate states that lie between the normal and low light.}
For high-level adaptation, we use multi-task self-supervised learning to close the feature distance between \wwj{the intermediate states built by low-level adaptation.}
By combining low-level and high-level adaptation, we outperform state-of-the-art face detection methods even though we do not use the labels of dark faces. Our contributions are summarized as follows:

\begin{itemize}
  \item We propose a framework for dark face detection without annotated dark data. Through a joint low-level and high-level adaptation, our model achieves superior performance compared with state-of-the-art face detection and adaptation methods.
  \item For low-level adaptation, we design a bidirectional scheme. \wwj{Through brightening low light data and distorting normal light data with noise and color bias}, we set up intermediate states and make two domains each take a step towards each other.
  \item For high-level adaptation, we introduce cross-domain self-supervised learning for feature adaptation. With context-based and contrastive learning, we comprehensively close the feature distance among multiple domains and further strengthen the representation.
\end{itemize}

\section{Related Works}
\label{sec:related_works}
{\flushleft {\bf Low Light Enhancement.}} 
Low illumination is a common kind of visual distortion, which might be caused by undesirable shooting conditions, wrong camera operations, and equipment malfunctions, \etc
There have been many literatures for low light enhancement.
Histogram equalization and its variants~\cite{Enhance_HE} stretch the dynamic range of the images.
Dehazing-based methods~\cite{Enhance_Dehaze} regard dark images as inverted hazy images.
Retinex theory assumes that images can be decomposed into illumination and reflectance.
Based on the Retinex theory, a large portion of works~\cite{Enhance_LIME,Enhance_MF} estimate illumination and reflectance, then process each component separately or simultaneously.
Recent methods are mainly based on deep learning.
Some design end-to-end processing models~\cite{Enhance_ZeroDCE}, while
some inject traditional ideas such as the Retinex theory~\cite{Enhance_SICE,Enhance_RetinexNet,Enhance_KinD}.
Besides processing 8-bit RGB images, there are also models for RAW images~\cite{Enhance_SID}.

The problem is that these methods are mainly designed for human vision rather than machine vision.
How pixel-level adjustment can benefit and guide high-level tasks has not been well explored.
In this paper, we provide corresponding solutions for dark face detection.

{\flushleft {\bf Face Detection.}} 
Early face detectors rely on hand-crafted features~\cite{FD_IJCV04}, which are now replaced by deep features learned from data-driven convolutional neural networks.
Inherit from generic object detection, typical face detectors can be classified into two categories: two-stage and one-stage.
Two-stage models~\cite{FD_FasterRCNN_NIPS15,FD_FasterRCNN_TPAMI17} first generate region proposals, then refine them for the final detection.
One-stage models~\cite{FA_Focal_ICCV17} instead directly predict the bounding boxes and confidence.
The difference between generic object and face detection is that, in face detection, the scale variation is often much larger.
Existing methods solve this problem by multi-scale image and feature pyramids~\cite{FA_FTF_CVPR17,FA_FPN_CVPR17}, or various anchor sampling and matching strategies~\cite{FA_SFRAP_CVPR18,FA_S3FD_ICCV17,FA_GSSIFD_CVPR19}.

Despite the prosperity of face detection researches, existing models seldom consider the scenario of insufficient illumination. 
In this paper, we propose a dark face detector that outperforms state-of-the-art methods even without using dark annotations.

\begin{figure*}[t]
    \centering
  \includegraphics[width=0.99\linewidth]{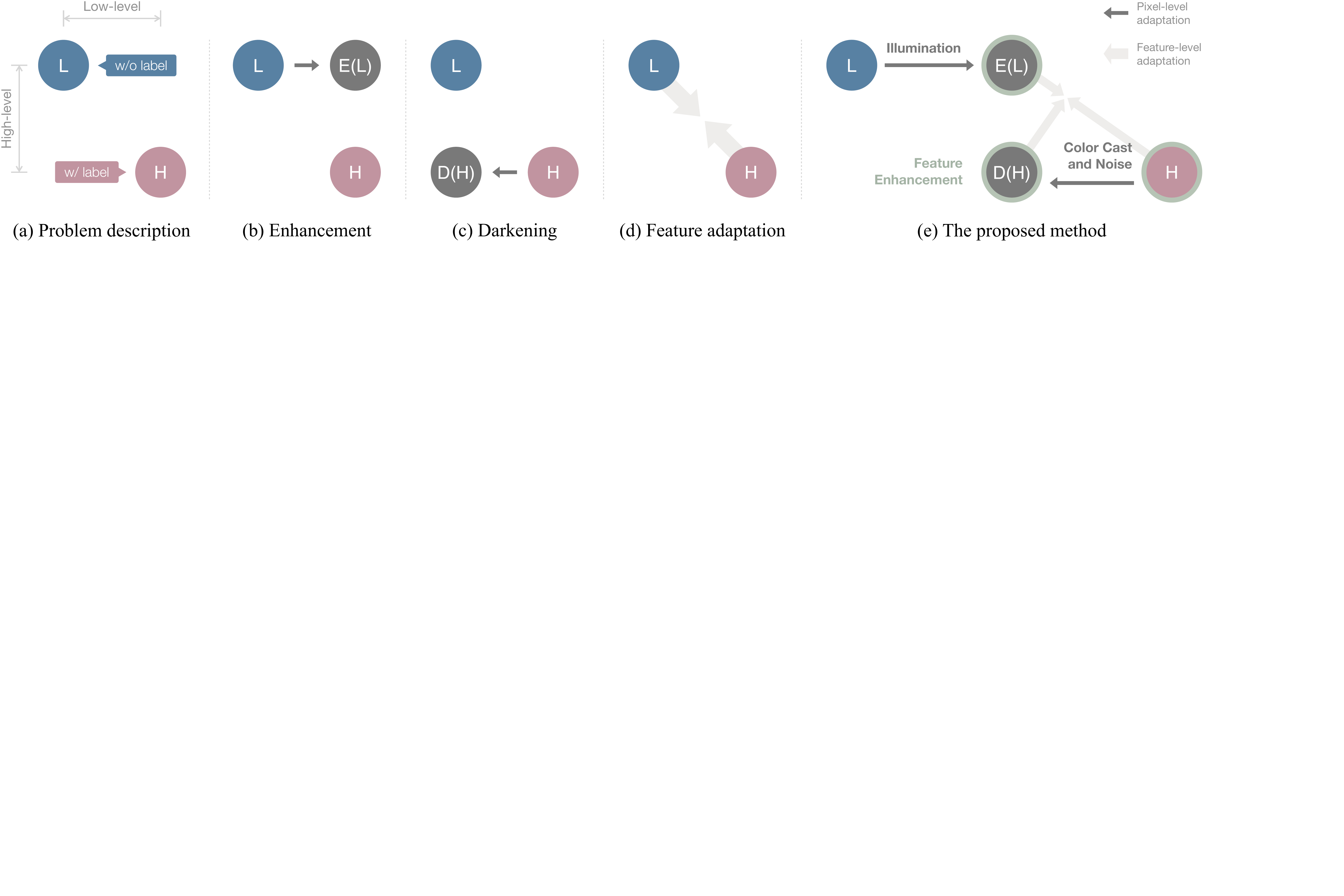}
  \caption{Comparison of different adaptive low light detection techniques. $L$: low light data. $H$: normal light data. Existing enhancement-based, darkening-based, and feature adaptation methods either ignore the high-level gap, or have limited effects due to the huge and complex gap between $L$ and $H$. Our method instead considers both low-level and high-level adaptation, therefore achieves better performance.}
    \label{fig:motivation}
\end{figure*}

{\flushleft {\bf Dark Object Detection.}} 
With the rapid development of deep learning, object detection has attracted more and more attention.
However, few efforts have been made for dark objects.
For RAW images, YOLO-in-the-Dark~\cite{FA_YOLODARK_ECCV20} merges models pre-trained in different domains using glue layers and a generative model.
For RGB images, Loh \etal build the ExDark~\cite{FA_ExDark_CVIU19} dataset and analyze the low light images using both hand-crafted and learned features.
DARK FACE~\cite{DARKFACE} is a large-scale low light face dataset, giving birth to a series of dark face detectors in the UG$^{2}$ Prize Challenge\footnote{http://cvpr2020.ug2challenge.org/}.
However, most of these models highly rely on annotations, thus are of limited flexibility and robustness.

To get rid of the dependency on labels, Unsupervised Domain Adaptation (UDA) may be a plain solution~\cite{DA_Diversify_CVPR19,DA_Cross_CVPR18}.
Although UDA has been demonstrated to be effective in many applications, due to the huge gap between normal and low light, these methods have limited performance in dark face detection.
In this paper, we propose a superior method by combining low- and high-level adaptation.

\begin{figure}[t]
    \centering
  \includegraphics[width=\linewidth]{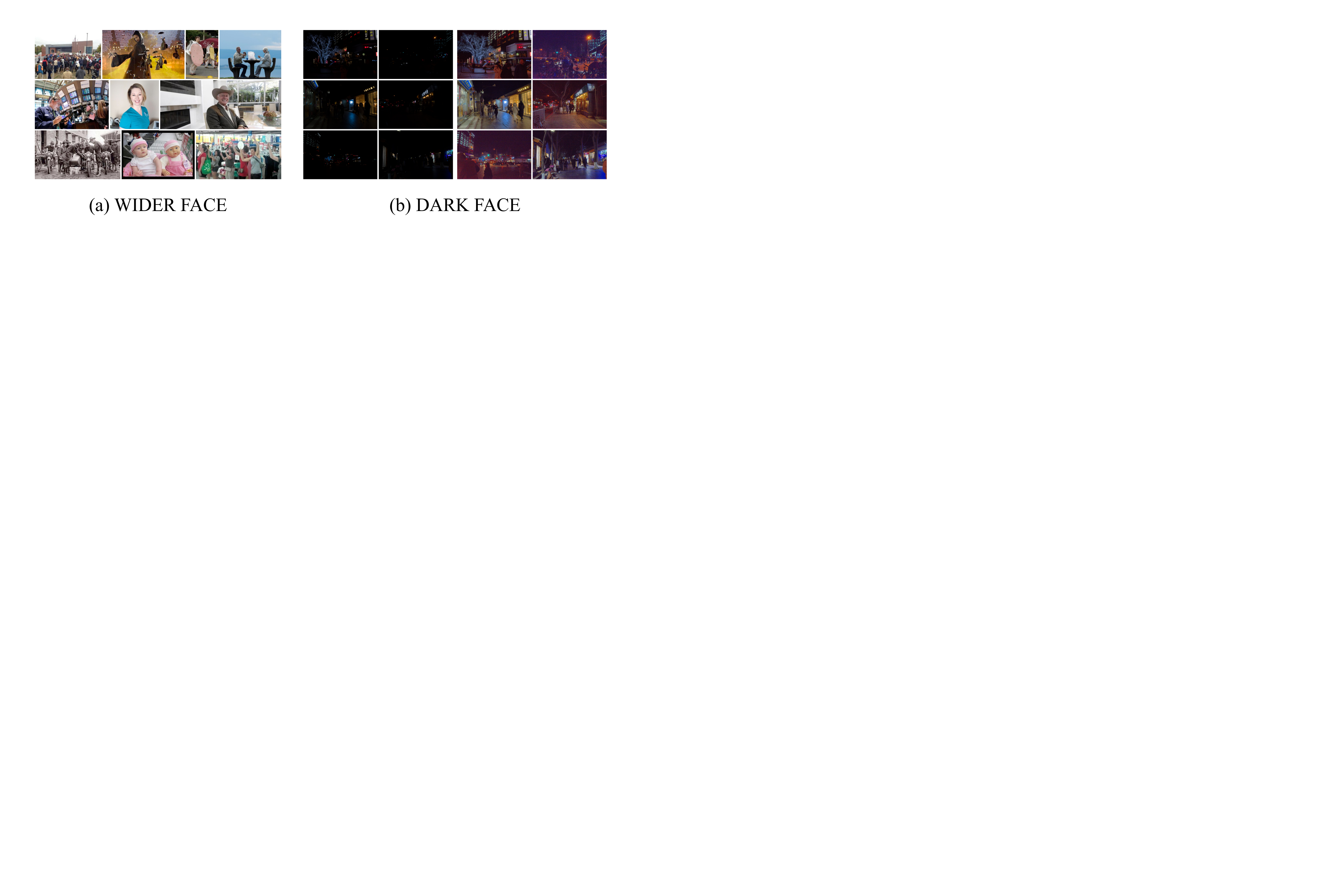}
  \caption{Comparison of WIDER FACE and DARK FACE. On the right, DARK FACE is enhanced for better visibility.}
    \label{fig:data_comparison}
\end{figure}

\begin{figure*}[t]
    \centering
  \includegraphics[width=\linewidth]{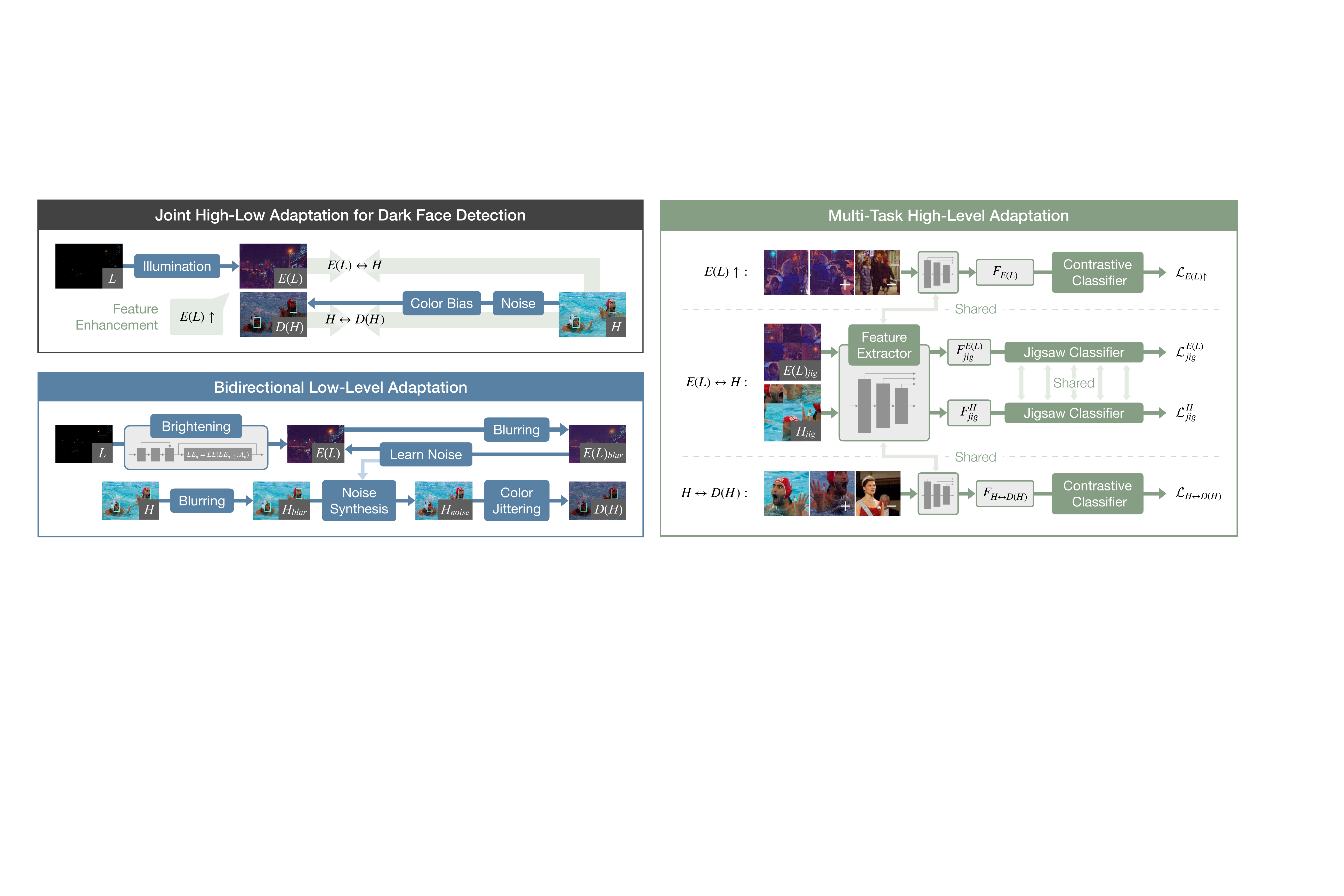}
  \caption{The overview of our joint High-Low Adaptation (HLA) framework for dark face detection. Low-level adaptation fills the gap by creating intermediate states. We bidirectionally brighten the low light data as well as distort the normal light data with noise and color bias. Based on the built intermediate states, we use multi-task cross-domain self-supervised learning to fill the high-level gap.}
    \label{fig:framework}
\end{figure*}

\begin{figure}[t]
    \centering
  \includegraphics[width=\linewidth]{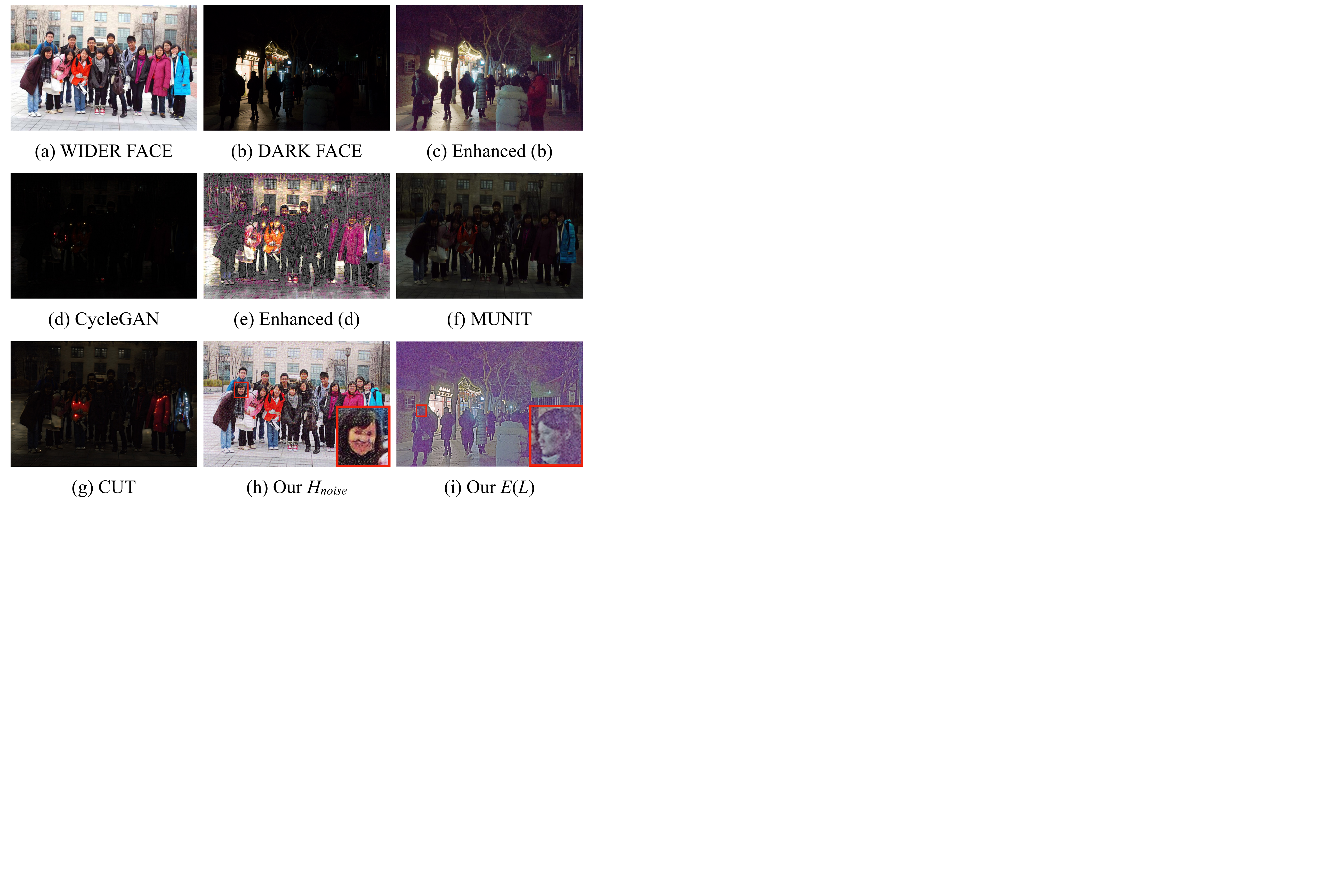}
  \caption{Results of transferring between WIDER FACE and DARK FACE. (b) and (d) are enhanced for better visualization.}
    \label{fig:drk_comparison}
\end{figure}

\section{Joint Adaptation for Dark Face Detection}

In this section, we firstly introduce the motivation of our learning paradigm, then describe the detailed designs.

\subsection{Motivation}
\label{sec:motivation}

The task is to adapt face detectors trained on normal light data $H$ to unlabeled low light data $L$. 
As shown in Fig.~\ref{fig:motivation}, existing methods can be roughly divided into three categories: enhancement, darkening, and feature adaptation.
\textbf{Enhancement}-based methods~\cite{Motivaiton_Enh_ChoBKAP20} brighten the low light images and directly test on them.
They usually require no model fine-tuning, therefore are highly flexible. 
\textbf{Darkening}-based methods~\cite{Moti_drk_LeeRK20,Moti_drk_ICCV19,Moti_drk_IJCNN} first darken the normal light data into a dark version, then re-train the model on the transferred annotated data.
Enhancement and darkening are all pixel-level.
For \textbf{feature adaptation}, typical methods use alignment~\cite{DA_FA_ECCV16}, adversarial learning~\cite{DA_AL_JMLR16}, or pseudo labeling~\cite{DA_Cross_CVPR18} to directly adapt the features of the model.

The problem for dark face detection is that the gap between $H$ and $L$ is too huge and complex for existing methods to handle.
As shown in Fig.~\ref{fig:data_comparison}, the images in WIDER FACE~\cite{WIDERFACE} and DARK FACE~\cite{DARKFACE} not only have different pixel-level appearance (bright v.s. dark, clean v.s. noisy), but also contain different objects and scenes (photos, paintings v.s. street views).
However, enhancement- and darkening-based methods only consider the pixel-level gap.
Feature adaptation methods try to fill the whole gap in one step. But as shown Sec.~\ref{sec:comparison}, the effect is limited.


To jointly fill both pixel-level and feature-level gaps for dark face detection, we propose a High-Low Adaptation (HLA) scheme.
As shown in Fig.~\ref{fig:motivation}~(e), we set low-level intermediate states between $L$ and $H$, and based on these states adapt the corresponding high-level representations.
Specifically, the low-level distance is reduced by both \textit{enhancing} and \textit{darkening}.
Compared with $L$-to-$H$ or $H$-to-$L$ unidirectional translation, our bidirectional translation: $L$-to-$E(L)$ and $H$-to-$D(H)$, can not only ease the difficulty of adaptation, but also provide more tools for feature-level adaptation.
The high-level distance is reduced by pushing the feature spaces of multiple states towards each other.
Moreover, the feature representation is further enhanced by contrastive learning.
While testing, we first process the image by $E(\cdot)$, then apply the adapted face detector.

The framework detail is shown in Fig.~\ref{fig:framework}. 
In the following, we will respectively introduce the proposed low-level and high-level adaptation schemes.

\subsection{Bidirectional Low-Level Adaptation}
\label{sec:low_level}

The challenge of low-level adaptation lies in two aspects.
One is the co-existence of the high-level gap, which can confuse pixel-level transfer models.
For example, we show the effect of some methods for transferring $H$ to $L$ in Fig.~\ref{fig:drk_comparison}.
Different from WIDER FACE, DARK FACE contains many street lights, vehicle highlights, and signboards.
Accordingly, CUT~\cite{Darkening_CUT} generates weird lights on human bodies, and CycleGAN~\cite{Darkening_CycleGAN} generates street lights on faces.
MUNIT~\cite{Darkening_MUNIT} can distinguish content and style, therefore has no street light artifact.
However, MUNIT cannot completely darken the image, and the result is visually far from $L$.

The other challenge is the difficulty of low light enhancement itself.
Existing low light enhancement methods are mainly designed for human vision rather than machine vision.
Some methods draw black edges, keep noisy parts dark, or enhance the contrast to improve the comprehensive visual quality, which can damage the high-level detection performance.
Moreover, images in DARK FACE suffer from intensive noise and color bias.
However, existing denoising and color reconstruction methods are not robust enough to handle this extreme case.

To solve the above challenges, we propose the bidirectional low-level adaptation scheme.
Low light degradation is a complex process. We roughly decompose the related factors into three aspects: illumination, noise, and color bias.
Although denoising and color correction is difficult, adding noise and applying color bias inversely is relatively easy.
On this basis, we bright $L$ into $E(L)$, and distort $H$ with noises and color bias to form $D(H)$.
Compared with $L$ and $H$, $E(L)$ and $D(H)$ are more similar.
In this way, we ease the difficulty of adaptation by making $L$ and $H$ take a step towards each other.
Also, by formulating the specific components of low light degradation, the transfer model will not be disturbed by the semantic gap between the domains.
In the following, we will introduce the detailed designs of each procedure.

{\flushleft {\bf Brightening.}} 
Different from common low light enhancement tasks, here we want to adjust the illumination without denoising or color reconstruction.
Moreover, low light images suffer from nonuniform illumination.
Some faces may be brightened by street lights, while some may be covered in severe darkness.
Therefore, we also need to prevent over-exposure as well as under-exposure.

Our module is based on nonlinear curve mapping~\cite{Enhance_ZeroDCE}, which is made up of iterative quadratic curves $LE(\cdot)$:
\begin{align}
\label{eq:zero_dce}
    LE(x,A) & = x + Ax(1-x), \\
    LE_{n} & = LE(LE_{n-1};A_n),
\end{align}
where $LE_{0}$ is the input image, $LE_{n}$ is the result at iteration $n$, and $A_n$ is a pixel-wise three-channel adjustment map estimated by neural networks.
Compared with common end-to-end or Retinex-based deep enhancement methods, curve mapping does not introduce extra noise or artifacts.
We follow \cite{Enhance_ZeroDCE} to use a 7-layer CNN with symmetrical skip-connections and the corresponding training objectives.

\begin{figure}[t]
    \centering
  \includegraphics[width=\linewidth]{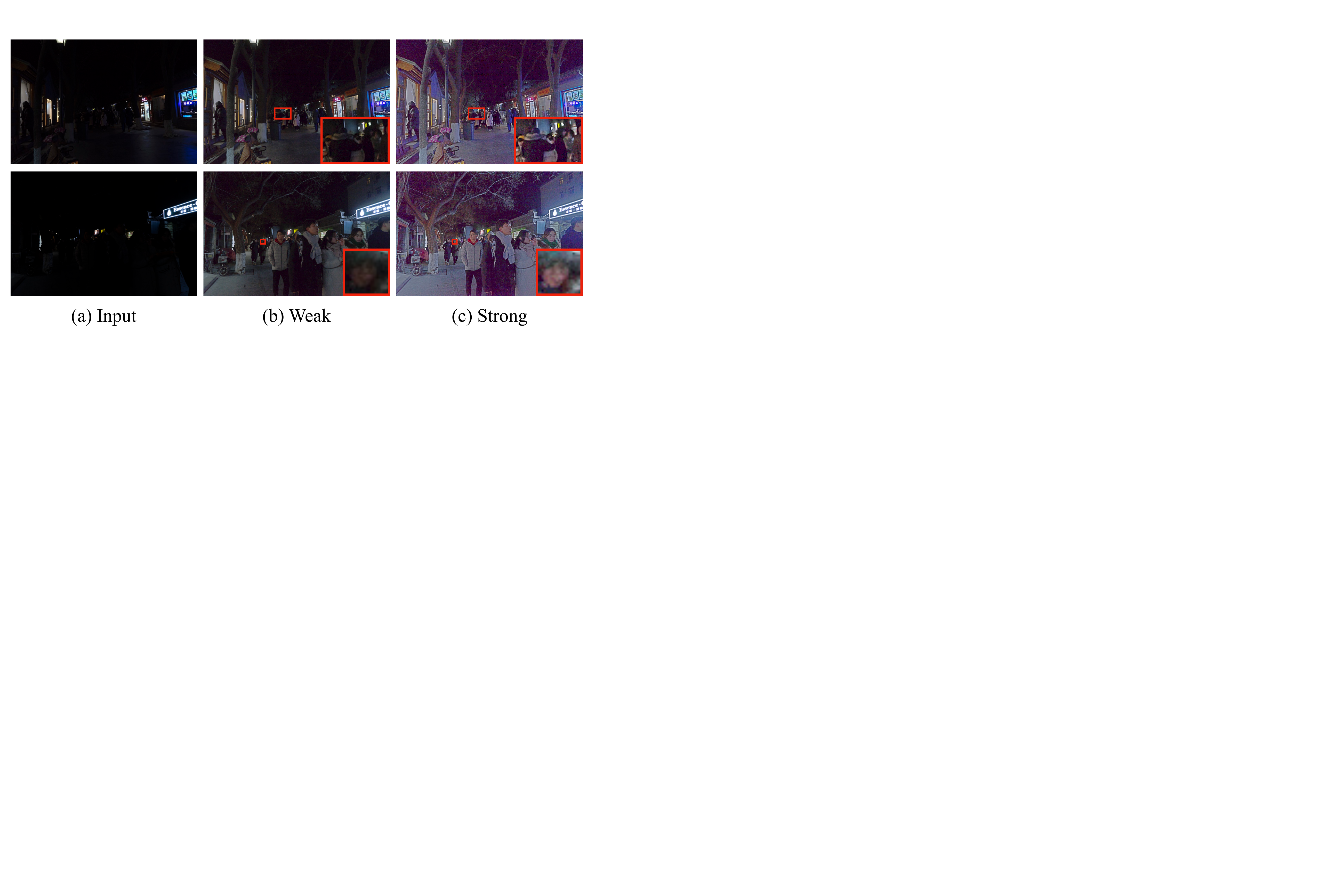}
  \caption{Effects of weak and strong brightening. Compared with (c), many faces are still covered in darkness in (b).}
    \label{fig:enh_comparison}
\end{figure}

The issue of \cite{Enhance_ZeroDCE} is that, the enhancement is conservative (Weak).
As shown in Fig.~\ref{fig:enh_comparison}~(b), many faces are still covered in darkness.
This is because further enhancing the image can bring more noise, and \cite{Enhance_ZeroDCE} choose to hide these noises in darkness, so that the visual quality of the whole image is better.
We instead propose strong illumination enhancement (Strong).
By doubling the iteration number in Eq.~(\ref{eq:zero_dce}) and widening the curve estimation network, the model can enhance the image with higher brightness.
The drawback may be that noise and color bias come along, but we can leave it to the following $H \rightarrow D(H)$ process.
This is also the difference between our enhancement module and common low light enhancement methods.

{\flushleft {\bf Noise Synthesis.}}
Although the pixel-level distance can be reduced by brightening, the gap between $E(L)$ and $H$ is still challenging.
Therefore, we further decompose the gap remained into color and noise.
Also, by separating out the color, we can use color to guide the noise synthesis process.

As shown in Fig.~\ref{fig:framework}, we first blur $E(L)$ by a strong Bilateral filter of $d=25$ and $\sigma=75$.
The blurring result $E(L)_{blur}$ works as the color guidance.
Then, a Pix2Pix~\cite{Pix2Pix} is trained for transferring from $E(L)_{blur}$ to $E(L)$.
Finally, we blur $H$ in the same way and use the trained Pix2Pix to add noise.
As shown in Fig.~\ref{fig:drk_comparison}, $H_{noise}$ successfully imitates the noise pattern of $E(L)$.
Their difference in color distribution will be handled in the next step.

{\flushleft {\bf Color Jittering.}}
We want the color distribution of $D(H)$ to match that of $E(L)$.
Based on statistical analysis, we set the jittering range to brightness: (0.4, 1.2), contrast: (0.6, 1.4), saturation: (0.6, 1.4) and hue: (0.8, 1.2).


\subsection{Multi-Task High-Level Adaptation}
\label{sec:high_level}

Most feature adaptation methods are based on alignment, pseudo labeling, and adversarial learning.
However, alignment and pseudo labeling cannot well handle the huge gaps, while adversarial learning is not stable.
We instead fully use the natural information of the images themselves, \ie, self-supervised learning.
By forcing the self-supervised learning classifiers to be shared across domains, the features are forced to be mapped into the same high dimensional subspace, therefore closing the high-level gap.

To push $E(L)$, $H$ and $D(H)$ towards each other, we first close $E(L)$-$H$ by cross-domain context-based self-supervised learning, then close $H$-$D(H)$ by cross-domain contrastive learning.
We further enhance the representation of $E(L)$ by single-domain contrastive learning.
The whole adaptation works in a multi-task way.
In the following, we will introduce the details of each learning scheme.

{\flushleft {\bf Closing E(L) and H.}}
Context-based self-supervised learning designs pretext tasks, through which the model can learn to understand the spatial context of objects.
Here, we use the jigsaw puzzling game~\cite{JIGSAW}. 
We have also tried rotation~\cite{ROTATION} and combining jigsaw with rotation, but find that using jigsaw alone works the best.
One possible explanation for this may be that many images in WIDER FACE are paintings or advertisements, where the faces may have strange angles.
Therefore, the rotation prediction pretext task can be ambiguous.

Similar to \cite{JIGSAW_DG}, we assemble $3 \times 3$ patches into a whole image and set the patch permutation number to $30$, \ie, 30 classification problem.
Denote $p_{jig}$ as the permutation label, and $\mathcal{L}_c$ as the cross-entropy loss, we have:
\begin{align}
    \mathcal{L}^{E(L)}_{jig} & = \mathcal{L}_{c}(F^{E(L)}_{jig}, p^{E(L)}_{jig}), \\
    \mathcal{L}^{H}_{jig} & = \mathcal{L}_{c}(F^{H}_{jig}, p^{H}_{jig}),
\end{align}
where $F_{jig}$ stands for the feature extracted from the corresponding domain.
$E(L)$ and $H$ share classification heads, which can force the semantic features to be mapped into the same space, therefore closing high-level gaps.
The final loss for closing $E(L)$ and $H$ is:
\begin{align}
    \mathcal{L}_{E(L) \leftrightarrow H} = \mathcal{L}^{E(L)}_{jig}  + \mathcal{L}^{H}_{jig}.
\end{align}

{\flushleft {\bf Closing H and D(H).}}
The idea of contrastive learning is that, given a query $v$, identifying its ``positive'' pair $v^{+}$ and ``negatives'' pairs $v^{-} = \{v^{-}_1, v^{-}_2 ..., v^{-}_N\}$.
With similarity measured by dot product, the objective $\mathcal{L}_{q}(v,v^{+},v^{-})$ is:
\begin{align}
\mathcal{L}_{q} & = -\text{log} \left [ \frac {\sigma(v,v^{+})}
									{\sigma(v,v^{+}) + \sum^{N}_{n=1} \sigma(v,v^{-}_n)} \right ], \\
\sigma(x,y) & = \text{exp}(x \cdot y / \tau),
\end{align}
where $\tau$ is a temperature hyper-parameter.
Intuitively, this is an $(N+1)$ classification problem.

To reduce the distance between $H$ and $D(H)$, we take advantage of the behavior that contrastive learning brings positive samples closer.
In specific, we make the positive pair of $H$ to be the patch from $D(H)$, and vice versa:
\begin{align}
\tilde{\mathcal{L}}_{H \leftrightarrow D(H)} & = \mathcal{L}_q	(H, D(H)^{+}, H^{-}) \nonumber \\ 
				   & + \mathcal{L}_q	(D(H), H^{+}, D(H)^{-}).
\end{align}
In this way, the feature similarity between $H$ and $D(H)$ can be improved, and the high-level gap can be closed.

We also introduce single-domain contrastive learning on $H$ and $D(H)$ themselves to make the features better.
In the implementation, the above four losses are simplified by regarding $D(\cdot)$ as a part of the augmentation:
\begin{align}
\mathcal{L}_{H \leftrightarrow D(H)} & = \mathcal{L}_q	(D_i^*(H), D_j^*(H)^{+}, D_k^*(H)^{-}),
\end{align}
where $D^*(H)$ has a 50\% probability of being $H$, and 50\% of being $D(H)$.
While training, we use the Momentum Contrast (MoCo)~\cite{MOCO} and follow \cite{MOCO_V2} for other settings.

{\flushleft {\bf Enhancing E(L).}}
We also find that it is beneficial to enhance the feature on $E(L)$ by contrastive learning:
 \begin{align}
    \mathcal{L}_{E(L)\uparrow} =\mathcal{L}_q	(E(L), E(L)^{+}, E(L)^{-}).
\end{align}

{\flushleft {\bf Final objective.}} Our model learns in a multi-task way.
Denote $\mathcal{L}_{det}$ as the detection loss, the final objective is:
 \begin{align}
    \mathcal{L} & = \lambda_{det} \mathcal{L}_{det}
    	    	      + \lambda_{E(L) \leftrightarrow H} \mathcal{L}_{E(L) \leftrightarrow H} \nonumber \\
		      & + \lambda_{H \leftrightarrow D(H)} \mathcal{L}_{H \leftrightarrow D(H)}
		      + \lambda_{E(L)\uparrow} \mathcal{L}_{E(L)\uparrow},
\end{align}
where $\lambda$s are hyper-parameters to balance different losses. 

\section{Experimental Results}

\subsection{Implementation Details}
\label{sec:implementation}

{\flushleft {\bf Network Architecture.}}
DSFD~\cite{DSFD} is used as the face detection baseline.
Our headers for self-supervised learning are added on the conv3\_3, conv4\_3, conv5\_3, conv\_fc7, conv6\_2, and conv7\_2 layers of the backbone. For more  details, please refer to the supplementary material.

{\flushleft {\bf Training and Evaluation Settings.}}
All experiments are based on WIDER FACE~\cite{WIDERFACE} and DARK FACE~\cite{DARKFACE}.
Our model is allowed to use the labels of WIDER FACE, but not allowed to use the labels of DARK FACE.
The framework is first pre-trained on WIDER FACE, then fine-tuned with both WIDER FACE and the images of DARK FACE.
Pre-training follows the same process of ~\cite{DSFD}.
For fine-tuning, the batch size is set to 8.
We use SGD with 0.9 momentum and 5e-4 weight decay.
The learning rate is set to 1e-4 for the first 20k iterations, and 1e-5 for another 40k iterations.
Fine-tuning takes about 15 hours with two GeForce RTX 2080Ti.
The testing process is the same as the original DSFD implementation.

For DARK FACE, we use the official train/test setting, and further split 500 images from the training set for validation.
Finally, there are 5500 images for training, 500 images for validation, and 4000 images for testing.
Performance is measured by mean Average Precision (mAP), and evaluated with the official tool\footnote{https://github.com/Ir1d/DARKFACE\_eval\_tools} of DARK FACE.

\begin{table}[t]
   \centering
   \small
   \caption{Comparison results on DARK FACE.}
   \label{table:comparison}
   \vspace{1mm}
    \begin{tabular}{l|l|c}
         \toprule
         Category & Method & mAP (\%) \\ 
         \midrule
	\multirow{8}{*}{Face  Detection}		& Faster-RCNN~\cite{Faster_RCNN} 		& \ \ 1.7 \\ 
	 							& SSH~\cite{SSH} 						& \ \ 6.9 \\ 
	 							& RetinaFace~\cite{RetinaFace}			& \ \ 8.6 \\ 
	 							& SRN~\cite{SRN} 						& \ \ 9.0 \\ 
	 							& SFA~\cite{SFA} 						& \ \ 9.3 \\ 
	 							& PyramidBox~\cite{PyramidBox}			& 12.5 \\ 
	 							& Small Hard Face~\cite{Small_Hard_Face} 	& 16.1 \\ 
	 							& DSFD~\cite{DSFD}					& 16.1 \\ 
	\midrule
	Enhancement					& Zero-DCE~\cite{Enhance_ZeroDCE}		& 37.7 \\ 
	(with Small Hard Face)			& MF~\cite{Enhance_MF}					& 38.3 \\ 
	\midrule
	\multirow{8}{*}{Enhancement}		& SICE~\cite{Enhance_SICE} 				& \ \ 4.7 \\ 
	\multirow{8}{*}{(with DSFD)}		& RetinexNet~\cite{Enhance_RetinexNet} 	& 12.0 \\ 
								& KinD~\cite{Enhance_KinD} 				& 15.8 \\ 
								& EnlightenGAN $\dagger$~\cite{Enhance_EnlightenGAN}	& 20.8 \\ 
								& EnlightenGAN~\cite{Enhance_EnlightenGAN} & 31.3 \\ 
								& Zero-DCE $\dagger$~\cite{Enhance_ZeroDCE}	& 37.3 \\ 
								& LIME~\cite{Enhance_LIME} 				& 40.7 \\ 
								& Zero-DCE~\cite{Enhance_ZeroDCE}		& 41.3 \\ 
								& MF~\cite{Enhance_MF}					& 41.4 \\ 
	\midrule
	\multirow{2}{*}{Darkening}			& MUNIT~\cite{Darkening_MUNIT}			& 29.7 \\ 
	\multirow{2}{*}{(with DSFD)}		& CycleGAN~\cite{Darkening_CycleGAN}		& 31.9 \\ 
								& CUT~\cite{Darkening_CUT}				& 32.7 \\ 
	\midrule
	\multirow{2}{*}{Unsupervised DA}	& OSHOT~\cite{DA_DInnocenteBBCT_ECCV20} 	& 25.4 \\ 
	\multirow{2}{*}{(with DSFD)} 		& Progressive DA~\cite{DA_WACV20}		& 28.5 \\ 
								& Pseudo Labeling~\cite{DA_Cross_CVPR18}	& 35.1 \\ 
	\midrule
	Fully Supervised				& Fine-tuned DSFD~\cite{DSFD}			& 46.0 \\ 
	\midrule
								& \textbf{Ours}			& \textbf{44.4} \\
           \bottomrule
    \end{tabular}
  \begin{tablenotes}
  	\footnotesize
  	\item[] $\dagger$ denotes retrained with DARKFACE.
  \end{tablenotes}
\end{table}

\begin{figure*}[t]
    \centering
  \includegraphics[width=\linewidth]{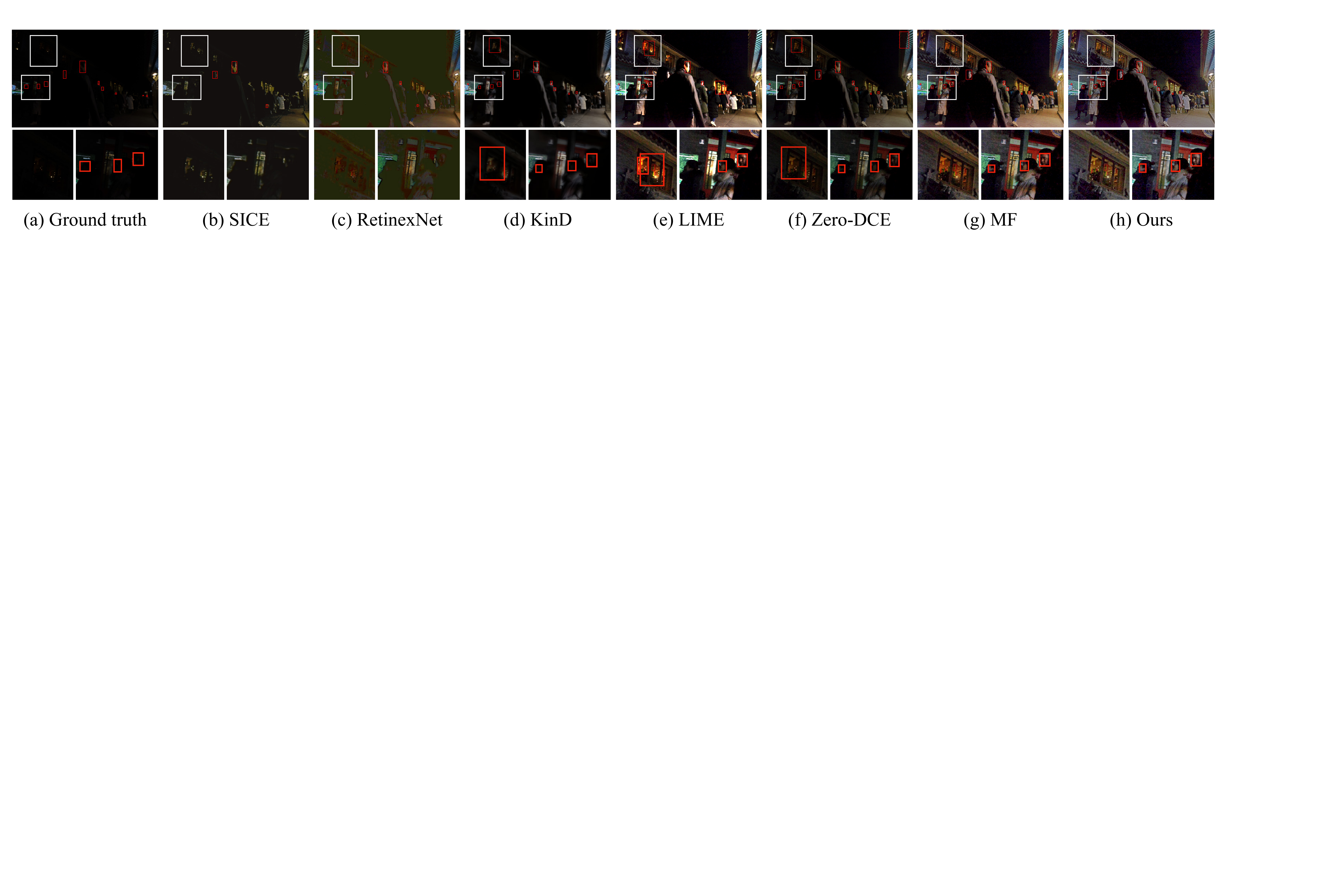}
  \caption{Qualitative comparison of different enhancement-based methods. (a) Input low light image and the ground truth boxes. (b)-(g) Results of low-light enhancement methods with DSFD. (h) Our result. }
    \label{fig:enh_comp_big}
\end{figure*}

\begin{figure}[t]
    \centering
  \includegraphics[width=0.95\linewidth]{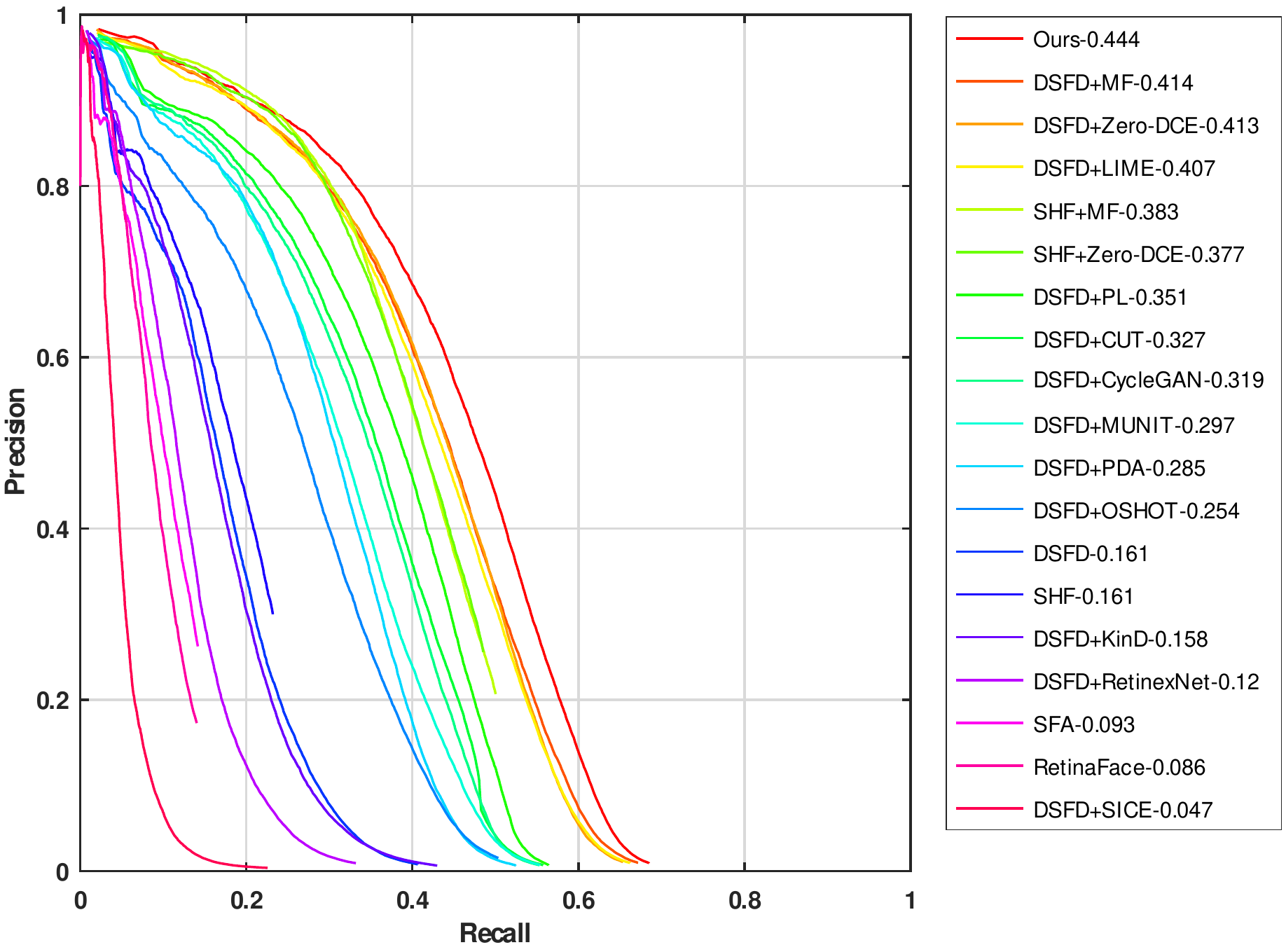}
  \caption{Precision-Recall (PR) curves on DARK FACE.}
  \label{fig:pr_curve}
\end{figure}

\subsection{Comparisons with State-of-the-Art Methods}
\label{sec:comparison}

The proposed model is compared with 22 state-of-the-art methods, covering the categories of face detection, low light enhancement, image-to-image translation, and unsupervised domain adaptation.
The benchmarking results are shown in Table~\ref{table:comparison} and Fig.~\ref{fig:pr_curve}.

{\flushleft {\bf Face Detection.}}
Our model is compared with seven face detectors and one generic object detector.
Due to the poor visibility caused by low light conditions, existing detectors all achieve undesirable performance.
As shown in Table~\ref{table:comparison}, Faster-RCNN\footnote{https://github.com/playerkk/face-py-faster-rcnn}~\cite{Faster_RCNN} (re-trained on WIDER FACE) performs worse than detection models designed especially for faces.
State-of-the-art face detection methods, SSH~\cite{SSH}, RetinaFace~\cite{RetinaFace}, SRN~\cite{SRN}, SFA~\cite{SFA}, and PyramidBox~\cite{PyramidBox}	 all have mAP scores less then 15\%, showing that insufficient illumination can greatly hurt the performance of high-level tasks.							
The best results here are achieved by DSFD~\cite{DSFD} and Small Hard Faces~\cite{Small_Hard_Face}, but their mAP scores are still unsatisfactory.
By adapting to the dark environment, our model outperforms these detectors by a significant margin.

{\flushleft {\bf Enhancement.}}
We also explore the effect of illumination adjustment, \ie, the scheme in Fig.~\ref{fig:motivation} (b).
We first use low light enhancement methods to enhance the DARK FACE images, then apply the face detectors.
Although DSFD and Small Hard Face are comparable on original dark images, when the images are brightened, DSFD outperforms Small Hard Face by 3.35\% in mAP on average.
This indicates that DSFD is of better robustness and generalization.
Therefore, in the rest of the experiments, we use DSFD as the baseline.
				
Although some low light enhancement methods can improve the performance to a large extent, some may even damage the detection performance.
This is because these methods introduce more artifacts to the images.
As shown in Fig.~\ref{fig:enh_comp_big}, SICE~\cite{Enhance_SICE} distorts the details. 
KinD~\cite{Enhance_KinD} over-denoises the images, leading to blurry edges and dull color.
RetinexNet~\cite{Enhance_LIME} instead generates weird green colors on dark regions.
These three methods widen the gap between the testing images and the daytime natural photographies, therefore hurt the performance of the face detector.
MF~\cite{Enhance_MF} and Zero-DCE~\cite{Enhance_ZeroDCE}, LIME~\cite{Enhance_LIME} can help DSFD better recognize faces.
The visual quality of their subjective results is also better.
However, compared with our model, their performance is still relatively poor. 
This is because when simply combining low light enhancement and face detection, the semantic gap between WIDER FACE and DARK FACE still remains.


{\flushleft {\bf Darkening.}}
Darkening-based adaptation, \ie, the scheme in Fig.~\ref{fig:motivation}~(c), proposes to re-train models on synthetic dark data.
Specifically, we first transfer WIDER FACE to DARK FACE, then use the transferred WIDER FACE to re-train DSFD. 
Most darkening-based methods~\cite{Moti_drk_ICCV19,Moti_drk_IV20} are based on the classic unsupervised image-to-image translation model CycleGAN~\cite{Darkening_CycleGAN}. We also test more powerful MUNIT~\cite{Darkening_MUNIT}, and the newest CUT~\cite{Darkening_CUT}.

Quantitative and qualitative results can be found in Fig.~\ref{fig:drk_comparison} and Table~\ref{table:comparison}, respectively.
Although MUNIT is more powerful than CycleGAN, the effect of benefitting dark face detection is worse.
This is because MUNIT cannot fully darken the image as shown in Fig.~\ref{fig:drk_comparison}~(f).
In Fig.~\ref{fig:drk_comparison}~(d) and (e), although the result of CycleGAN looks like night street views at the first glance, after enhancing the image, we can see that CycleGAN actually distorts the details and puts street light on faces.
CUT generates less artifact than CycleGAN, therefore the mAP score is higher.
However, compared with our model, the performances of darkening-based methods are all unsatisfactory.
This demonstrates our assumption that the gap between normal and low light is too huge and complex for pixel-level transfer models to handle.

\begin{table}[t]
   \centering
   \small
   \caption{Comparison with Top 10 teams (with labels) in the UG$^{2}$ Prize Challenge. Scores are copied from the official website.}
   \label{table:leaderboard}
   \vspace{1mm}
    \begin{tabular}{llccccc}
         \toprule
         Rank & Team Name &  mAP (\%) \\
         \midrule
1 & CAS-Newcastle-TUM & 62.45 \\
2 & CAS-NEU & 61.84 \\
\textbf{-}   & \textbf{Ours} & \textbf{44.44} \\
3 & MSFace & 42.71 \\
4 & iie & 40.49 \\
5 & NTU-MiRA & 37.50 \\
6 & DUTMedia & 35.65 \\
7 & SCUT-CVC & 35.18 \\
8 & IIAI VOS & 34.73 \\
9 & USTC-NELSLIP & 32.81 \\
10 & PHI-AI & 29.95 \\
           \bottomrule
    \end{tabular}
\end{table}

{\flushleft {\bf Unsupervised Domain Adaptation.}}
Most UDA methods are based on Faster-RCNN, which performs too poor on face detection as shown in Table~\ref{table:comparison}.
For a fair comparison, we re-implement all compared UDA methods with DSFD.

OSHOT~\cite{DA_DInnocenteBBCT_ECCV20} directly closes the gap by self-supervised learning of rotation angle prediction.
It is originally designed for one-shot adaptation. We change it into fine-tuning on the whole DARK FACE.
The performance of OSHOT is poor. This is because the gap between normal light and low light faces is too huge to be handled by feature adaptation.
Pseudo Labeling~\cite{DA_Cross_CVPR18} is a two-step progressive UDA method. 
It first uses CycleGAN to artificially generate training data, then uses pseudo labels to fine-tune the detector.
Compared with directly training on images synthesized by CycleGAN, the performance improves from 31.9\% to 35.1\% in mAP, demonstrating the effectiveness of pseudo labels. However, the mAP is still less than 40\%.
Progressive DA~\cite{DA_WACV20}	combines pixel-level transferring and feature-level adversarial learning. But adversarial learning still cannot close the huge gap between normal and low light domains.

{\flushleft {\bf With Dark Annotations.}}
Our model is also compared with face detection methods that have access to the labels of DARK FACE.
The result of fine-tuning DSFD with labels is shown in Table~\ref{table:comparison}.
We can see that our model is much closer to the supervised learning upper bound 46.0\% in mAP, demonstrating the effectiveness of our adaptation framework.
We also show the leader board of the UG$^{2}$ Prize Challenge\footnote{\ http://cvpr2020.ug2challenge.org/program19/leaderboard19\_t2.html} in Table~\ref{table:leaderboard}, where our model outperforms most of the teams.
Notice that the teams in UG$^{2}$ are allowed to use labels for training, while our model uses no DARK FACE annotations.

\begin{table}[t]
   \centering
   \small
   \caption{Ablation study results on DARK FACE. $\dagger$ denotes using the pyramid multi-scale testing scheme in DSFD.}
   \vspace{1mm}
   \label{table:aba_self_supervised}
    \begin{tabular}{l|l|l|l|c}
         \toprule
         $E(\cdot)$ 	& $E(L) \leftrightarrow H$ 	& $H \leftrightarrow D(H) $ & $E(L)\uparrow$ &  mAP (\%) \\ 
         \midrule
         -			& -			& -			& -	 		& 15.3 \\ 
         \midrule
	Weak 		& - 		& - 				& -	  		& 38.3 \\ 
	Strong		& - 		& - 				& -	  		& 39.1 \\ 
         \midrule
	-			& Rotation 	& - 			& -	  		& 22.7 \\ 
	-			& Jigsaw 		& - 			& -	  		& 26.9 \\ 
	-			& Rot + Jig 	& - 			& -	  		& 25.3 \\ 
         \midrule
	Strong		& - 		& Pseudo labels 	& -			& 40.2 \\ 
	Strong		& - 		& $H$ only  		& -			& 38.2 \\ 
	Strong		& - 		& Cross-domain 	& -			& 40.9 \\ 
	 \midrule
	Strong		& Jigsaw 	& - 				& -	  			& 40.2 \\ 
	Strong		& - 		& Cross-domain 	& $\checkmark$	& 41.1 \\ 
	Strong		& Jigsaw 	& Cross-domain  	& $\checkmark$	& 41.4 \\ 
	Strong		& Jigsaw 	& Cross-domain  	& $\checkmark$	& 44.4 $\dagger$ \\ 
           \bottomrule
    \end{tabular}
\end{table}

\subsection{Ablation Studies}
\label{sec:ablation}

To support our motivation and the joint high-low adaptation framework, in this section, we analyze the effect of each technical design.
The results are shown in Table~\ref{table:aba_self_supervised}.

{\flushleft {\bf Effectiveness of E(L).}}
Enhancing the testing images can improve the performance from 15.3\% to 39.1\% in mAP.
Compared with the baseline (Weak), the performance of our $E(L)$ is higher by 0.8\%, supporting our proposed strong enhancement.

{\flushleft {\bf Effectiveness of E(L) $\leftrightarrow$ H.}}
We show the effect of different choices of context-based self-supervised learning for closing $E(L)$ and $H$.
Using jigsaw alone works the best. Adding rotation can damage the performance.
As we mentioned in Sec.~\ref{sec:high_level}, since many images in WIDER FACE are paintings or advertisements, the rotation angle prediction pretext task can be ambiguous.
As shown in Table~\ref{table:jig_rot_acc}, the rotation top-1 classification accuracy on WIDER FACE is only slightly over random guess.
In comparison, although the jigsaw pretext task is a 30-class problem, the top-1 accuracies are higher than 85\%.
We also notice that for both jigsaw and rotation, the performance on DARK FACE is higher than that on WIDER FACE.
This is because the images in WIDER FACE are more diverse.

{\flushleft {\bf Effectiveness of H $\leftrightarrow$ D(H).}}
We further explore the strategy for closing $H$ and $D(H)$.
The proposed cross-domain contrastive learning scheme $\mathcal{L}_{H \leftrightarrow D(H)}$ can improve the mAP score by 1.8\%.
If we only use contrastive learning on $H$, the performance even drops from 39.1\% to 38.2\%.
This is because if we enhance features only in $H$, the detection model will concentrate more on $H$, therefore increasing the distance between $H$ and $E(L)$.
The result also supports our design of cross-domain contrastive learning and the necessity of setting the intermediate domain $D(H)$.

Contrastive learning can be regarded as a kind of ``soft'' label.
Naturally, we wonder about the effect of ``hard'' label.
We test the result of directly training with transferred labels on $D(H)$, \ie, pseudo labeling.
The mAP can be improved from 39.1\% to 40.2\%, but the improvement is smaller than using our contrastive learning.
This is because representation learning based on ``soft'' labels can avoid the inaccuracy of manual annotations and better refine the features.

\begin{table}[t]
   \centering
   \small
   \caption{Top-1 classification accuracy of jigsaw and rotation self-supervised learning pretext tasks in different domains.}
   \vspace{1mm}
   \label{table:jig_rot_acc}
    \begin{tabular}{lcccc}
         \toprule
         Layer 	& Jig, $E(L)$ & Jig, $H$ & Rot, $E(L)$ & Rot, $H$ \\ 
         \midrule
         conv3\_3 	& 97.5\% & 80.6\% & 19.2\% & 15.4\% \\
         conv4\_3 	& 98.9\% & 87.8\% & 33.6\% & 26.2\% \\
         conv5\_3 	& 99.0\% & 89.6\% & 51.2\% & 36.7\% \\
         conv\_fc7 	& 99.3\% & 89.7\% & 54.9\% & 33.5\% \\
         conv6\_2 	& 99.3\% & 89.7\% & 51.2\% & 28.3\% \\
         conv7\_2 	& 99.3\% & 90.0\% & 41.8\% & 19.3\% \\
         \midrule
         average 		& 98.9\% & 87.9\% & 42.0\% & 26.6\% \\
           \bottomrule
    \end{tabular}
\end{table}

{\flushleft {\bf Combination Effect.}}
Finally, we demonstrate the combination effect of our design.
Enhancing the feature on $E(L)$ ($\mathcal{L}_{E(L)\uparrow}$) can further improve the mAP score.
The full version of our model achieves the best performance, demonstrating the effectiveness of our joint high-level and low-level adaptation framework.

DSFD uses a pyramid multi-scale testing scheme.
Although it can improve the performance, the running time increases from 1.25 hours to 10 hours.
Even without this multi-scale scheme, \ie, using a more weak face detection baseline, the performance of our model (41.4\% in mAP) can still outperform most of the state-to-the-arts in Table~\ref{table:comparison}.

\section{Conclusion}

We design a joint high-level and low-level adaptation framework for dark face detection.
We propose a bidirectional pixel translation pipeline for the low level, and a multi-task adaptation strategy based on self-supervised learning for the high level.
Our framework demonstrates the potential of joint high-low adaptation and can inspire other related low light high-level vision tasks.

{\small
\bibliography{reference}
}

\end{document}